\documentclass{article}

 \PassOptionsToPackage{numbers, compress}{natbib}

\usepackage[preprint]{neurips_2019}

\usepackage[utf8]{inputenc} 
\usepackage[T1]{fontenc}    
\usepackage{hyperref}       
\usepackage{url}            
\usepackage{booktabs}       
\usepackage{amsfonts}       
\usepackage{nicefrac}       
\usepackage{microtype}      

\usepackage{graphicx}
\usepackage{amsmath}
\usepackage{bm}
\usepackage{subfigure}
\usepackage{wrapfig}

\title{Hashing based Answer Selection}

\author{
  Dong Xu \\
  National Key Lab. for Novel Software Tech.\\
  Dept. of Comp. Sci. and Tech.\\
  Nanjing University, Nanjing 210023, China \\
  xud@lamda.nju.edu.cn \\
  \And
  Wu-Jun Li \\
  National Key Lab. for Novel Software Tech.\\
  Dept. of Comp. Sci. and Tech.\\
  Nanjing University, Nanjing 210023, China \\
  liwujun@nju.edu.cn \\
  }

\begin{document}

\maketitle

\begin{abstract}
   Answer selection is an important subtask of question answering~(QA), where deep models usually achieve better performance. Most deep models adopt question-answer interaction mechanisms, such as attention, to get vector representations for answers. When these interaction based deep models are deployed for online prediction, the representations of all answers need to be recalculated for each question. This procedure is time-consuming for deep models with complex encoders like BERT which usually have better accuracy than simple encoders. One possible solution is to store the matrix representation~(encoder output) of each answer in memory to avoid recalculation. But this will bring large memory cost. In this paper, we propose a novel method, called \underline{h}ashing based \underline{a}nswer \underline{s}election~(HAS), to tackle this problem. \mbox{HAS} adopts a hashing strategy to learn a binary matrix representation for each answer, which can dramatically reduce the memory cost for storing the matrix representations of answers. Hence, HAS can adopt complex encoders like BERT in the model, but the online prediction of HAS is still fast with a low memory cost. Experimental results on three popular answer selection datasets show that \mbox{HAS} can outperform existing models to achieve state-of-the-art performance.
\end{abstract}

\section{Introduction} \label{sec:intro}

  Question answering~(QA) is an important but challenging task in natural language processing~(NLP) area. Answer selection~(answer ranking) is one of the key components in many kinds of QA applications, which aims to select the corresponding answer from a pool of candidate answers for a given question. For example, in community-based question answering~(CQA) tasks, all answers need to be ranked according to the quality. In frequently asked questions~(FAQ) tasks, the most related answers need to be returned back for answering the users' questions.

  One main challenge of answer selection is that both questions and answers are not long enough in most cases. As a result, they usually lack background information and knowledge beyond the context~\cite{deng2018knowledge}. This phenomenon limits the performance of answer selection models. Deep neural networks~(DNN) based models, also simply called deep models, can partly tackle this problem by using pre-trained word embeddings. Word embeddings pre-trained on language corpus contains some common knowledge and linguistic phenomena, which is helpful to select answers. Furthermore, deep models can automatically extract complex features, while shallow~(non-deep) models~\cite{cui2015question,yih2013question,tellez-valero2011learning,robertson1996okapi,wang2010probabilistic} typically need manually designed lexical features. Deep models have achieved promising performance for answer selection in recent years~\cite{tan2016lstm,santos2016attentive,tay2018cross,tran2018multihop,deng2018knowledge}. Hence, deep models have become more popular than shallow models for answer selection tasks.

  Most deep models for answer selection are constructed with similar frameworks which contain an encoding layer~(also called encoder) and a composition layer~(also called composition module). Traditional models usually adopt convolutional neural networks~(CNN)~\cite{feng2015applying} or recurrent neural networks~(RNN)~\cite{tan2016lstm,tran2018multihop} as encoders. Recently, complex pre-trained models such as \mbox{BERT}~\cite{devlin2018bert} and \mbox{GPT-2}~\cite{radford2019language}, are proposed for NLP tasks. \mbox{BERT} and \mbox{GPT-2} adopt \mbox{Transformer}~\cite{vaswani2017attention} as the key building block, which discards CNN and RNN entirely. \mbox{BERT} and \mbox{GPT-2} are typically pre-trained on a large-scale language corpus, which can encode abundant common knowledge into model parameters. This common knowledge is helpful when \mbox{BERT} or \mbox{GPT-2} is fine-tuned on other tasks.

  The output of the encoder for each sentence of either question or answer is usually represented as a matrix and each column or row of the matrix corresponds to a vector representation for a word in the sentence. Composition modules are used to generate vector representations for sentences from the corresponding matrices. Composition modules mainly include pooling and question-answer interaction mechanisms. Question-answer interaction mechanisms include attention~\cite{tan2016lstm}, attentive pooling~\cite{santos2016attentive}, multihop-attention~\cite{tran2018multihop} and so on. In general, question-answer interaction mechanisms have better performance than pooling. However, interaction mechanisms bring a problem that the vector representations of an answer are different with respect to different questions. When deep models with interaction mechanisms are deployed for online prediction, the representations of all answers need to be recalculated for each question. This procedure is time-consuming for deep models with complex encoders like BERT which usually have better accuracy than simple encoders. One possible solution is to store the matrix representation~(with float or double values) of each answer in memory to avoid recalculation. But this will bring large memory cost.
  
  In this paper, we propose a novel method, called \underline{h}ashing based \underline{a}nswer \underline{s}election~(HAS), to tackle this problem. The main contributions of HAS are briefly outlined as follows:
  \begin{itemize}
  		\item \mbox{HAS} adopts a hashing strategy to learn a binary matrix representation for each answer, which can dramatically reduce the memory cost for storing the matrix representations of answers. To the best of our knowledge, this is the first time to use hashing for memory reduction in answer selection.
  		\item By storing the (binary) matrix representations of answers in the memory, HAS can avoid recalculation for answer representations during online prediction. Subsequently, HAS can adopt complex encoders like BERT in the model, but the online prediction of HAS is still fast with a low memory cost.
		\item Experimental results on three popular answer selection datasets show that \mbox{HAS} can outperform existing models to achieve state-of-the-art performance.
  \end{itemize}
	

\section{Related Work}

\textbf{\emph{Answer Selection}} The earliest models for answer selection are shallow~(non-deep) models, which usually use bag-of-words~(BOW)~\cite{yih2013question}, manually designed rules~\cite{tellez-valero2011learning}, syntactic trees~\cite{wang2010probabilistic,cui2015question} as sentence features. Different upper structures are designed for modeling the similarity of questions and answers based on these features. The main drawback of shallow models are the lacking of semantic information by using only surface features. Deep models can capture more semantic information by distributed representations, which lead to better results than shallow models. Early deep models use pooling~\cite{feng2015applying} as the composition module to get vector representations for sentences from the encoder outputs which are represented as matrices. Pooling cannot model the interaction between questions and answers, which has been outperformed by new composition modules with question-answer interaction mechanisms. Attention~\cite{bahdanau2015neural} can generate a better representation of answers~\cite{tan2016lstm} than pooling, by introducing the information flow between questions and answers into models.  \cite{santos2016attentive} proposes attentive pooling for bidirectional attention. \cite{tran2018multihop} proposes a strategy of multihop attention which captures the complex relations between question-answer pairs. \cite{wan2016deep} focuses on the word by word similarity between questions and answers. \cite{wang2016inner} and \cite{chen2018rnn} propose inner attention which introduces the representation of question to the answer encoder through gates. \cite{tay2018cross} designs a cross temporal recurrent cell to model the interaction between questions and answers.

\textbf{\emph{BERT and Transfer Learning}} To tackle the problem of insufficient background information and knowledge in answer selection, one way is to introduce extra knowledge from other data. \cite{deng2018knowledge,min2017question,wiese2017neural} employ supervised transfer learning frameworks to pre-train a model from a source dataset. There are also some unsupervised transfer learning techniques~\cite{yu2018modelling,chung2018supervised}. BERT~\cite{devlin2018bert} is a recently proposed model for language understanding. Through training on a large language corpus, abundant common knowledge and linguistic phenomena can be encoded into the parameters. As a result, BERT can be transferred to a wide range of NLP tasks and has shown remarkable results.

\textbf{\emph{Hashing}} Hashing~\cite{li2016feature} tries to learn binary codes for data representations. Based on the binary code, hashing can be used to speedup retrieval and reduce memory cost. In this paper, we take hashing to reduce memory cost, by learning binary matrix representations for answers. There have already appeared many hashing techniques for learning binary representation~\cite{li2016feature,cao2017hashnet,hubara2016binarized}. To the best of our knowledge, there have not existed works to use hashing for memory reduction in answer selection.

\section{Hashing based Answer Selection}

In this section, we present the details of \underline{h}ashing based \underline{a}nswer \underline{s}election~(\mbox{HAS}), which can be used to solve the problem faced by existing deep models with question-answer interaction mechanisms. 

\begin{figure*}[htb]
	\centering
	\includegraphics[width=0.85\textwidth]{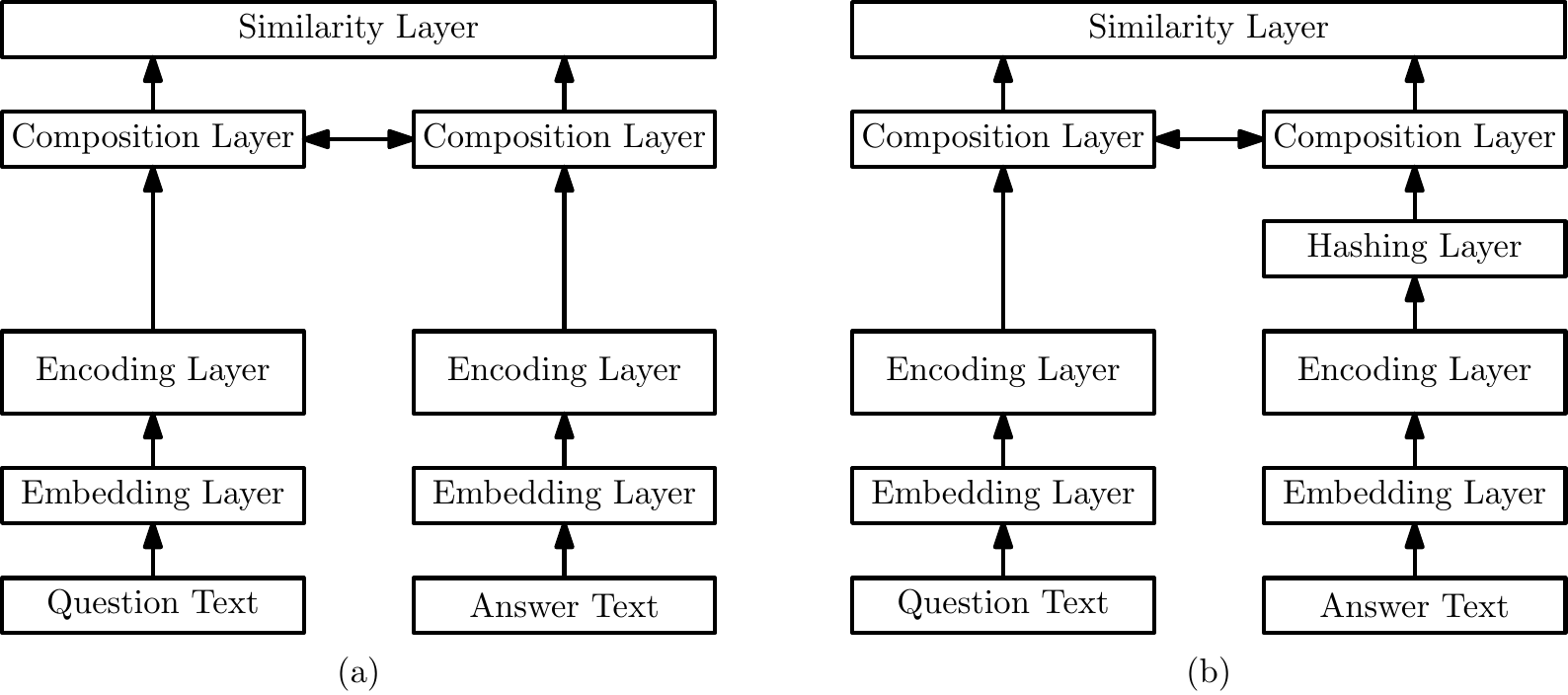}
	\vspace{-9pt}
	\caption{\label{model} (a) Framework of traditional deep models for answer selection. (b) Framework of HAS.}
	\vspace{-10pt}
\end{figure*}

The framework of most existing deep models is shown in Figure~\ref{model}(a). Compared with this framework, \mbox{HAS} has an additional \emph{hashing layer}, which is shown in Figure~\ref{model}(b). More specifically, \mbox{HAS} consists of an \emph{embedding layer}, an \emph{encoding layer}, a \emph{hashing layer}, a \emph{composition layer} and a \emph{similarity layer}. With different choices of encoders~(\emph{encoding layer}) and composition modules~(\emph{composition layer}) in HAS, several different models can be constructed. Hence, HAS provides a flexible framework for modeling. In the rest content of this section, we will introduce the layers of \mbox{HAS} in detail.

\subsection{Embedding Layer and Encoding Layer}
HAS is designed for modeling the similarity of question-answer pairs, thus the inputs to HAS are two sequences of words, corresponding to the question text and answer text respectively. Firstly, these sequences of words are represented by word embeddings through a word embedding layer. Suppose the dimension of word embedding is $E$, and the sequence length is $L$. The embeddings of question $q$ and answer $a$ are represented by matrices $\bm{Q} \in \mathbb{R}^{E \times L}$ and $\bm{A} \in \mathbb{R}^{E \times L}$ respectively. We use the same sequence length $L$ for simplicity. Then, these two embedding matrices $\bm{Q}$ and $\bm{A}$ are fed into an \emph{encoding layer} to get the contextual word representations. Different choices of \emph{embedding layers} and encoders can be adopted in HAS. Here, we directly use the \emph{embedding layer} and \emph{encoding layer} in BERT to utilize the common knowledge and linguistic phenomena encoded in BERT. Hence, the formulation of \emph{encoding layer} is as follows:
\begin{gather}
\bm{H}^q = BERT(\bm{Q}), \nonumber \\
\bm{H}^a = BERT(\bm{A}),  \nonumber
\end{gather}
where $\bm{H}^q, \bm{H}^a \in \mathbb{R}^{D \times L}$ are the contextual semantic features of words extracted by BERT, and $D$ is the output dimension of BERT.

\subsection{Hashing Layer}

The outputs of the \emph{encoding layer}~(encoder) for question $q$ and answer $a$ are $\bm{H}^q$ and $\bm{H}^a$, which are two real-valued~(float or double) matrices. When deep models with question-answer interaction mechanisms store the output of \emph{encoding layer}~($\bm{H}^a$) in memory to avoid recalculation, they will meet the high memory cost problem. For example, if we take float values for $\bm{H}^a$, the memory cost for only one answer $\bm{H}^a$ is $600$ KB when $L=200, D=768$. Here, $D=768$ is the output dimension of BERT. If the number of answers in candidate set is large, excessive memory cost will lead to impracticability, especially for mobile or embedded devices. 

In this paper, we adopt hashing to reduce memory cost by learning binary matrix representations for answers. More specifically, we take the sign function $y = sgn(x)$ to binarize the output of the \emph{encoding layer}. But the gradient of the sign function is zero for all nonzero inputs, which leads to a problem that the gradients cannot back-propagate correctly. $y = tanh(x)$ is a commonly used approximate function for $y = sgn(x)$, which can make the training process end-to-end with back-propagation~(BP). Here, we use a more flexible variant $y = tanh(\beta x)$ with a hyper-parameter $\beta \geq 1$. The derivative of $y = tanh(\beta x)$ is
\begin{gather}
\frac{\partial{y}}{\partial{x}} = \beta (1 - y^2). \nonumber
\end{gather}

By using this function, the formulation of \emph{hashing layer} is as follows:
\begin{gather}
\bm{B}^a = tanh(\beta \bm{H}^a),
\end{gather}
where $\bm{B}^a \in \mathbb{R}^{D \times L}$ is the output of \emph{hashing layer}. 

To make sure that the elements in $\bm{B}^a$ can concentrate to binary values $\mathbb{B} = \{\pm 1\}$, we add an extra constraint for this layer:
\begin{gather}
\mathcal{J}^c = || \bm{B}^a - \bm{\mathcal{B}}^a ||_F^2, \label{loss_c}
\end{gather}
where $\bm{\mathcal{B}}^a \in \mathbb{B}^{D \times L}$ is the binary matrix representation for answer $a$, $||*||_F$ is the Frobenius norm of a matrix. Here, $\bm{\mathcal{B}}^a$ is also a parameter to learn in HAS model. 

When the learned model is deployed for online prediction, the learned binary matrices for answers will be stored in memory to avoid recalculation. With binary representation, each element in the matrices only costs one bit of memory. Hence, the memory cost can be dramatically reduced.

\subsection{Composition Layer}
The outputs of \emph{encoding layer} and \emph{hashing layer} are matrices of size $D \times L$. Composition layers are used to compose these matrix representations into vectors. Pooling, attention~\cite{tan2016lstm}, attentive pooling~\cite{santos2016attentive} and other interaction mechanisms~\cite{tran2018multihop,wan2016deep} can be adopted in HAS. Interaction based modules usually have better performance than pooling based modules which have no question-answer interaction. Here, we take attention as an example to illustrate the advantage of HAS. More specifically, we adopt pooling for composing matrix representations of questions into question vectors, and adopt attention for composing matrix representations of answers into answer vectors. The formulation of the \emph{composition layer} is as follows:
\begin{align}
\bm{v}^q &= max\_pooling(\bm{H}^q), \nonumber \\
\bm{v}^a &= attention(\bm{B}^a, \bm{v}^q) = \sum_{i=1}^L \alpha_i \cdot \bm{b}^a_i, \nonumber \\
\alpha_i &\propto exp(\bm{m}^\top \cdot tanh(\bm{W}^a \cdot \bm{b}^a_i +  \bm{W}^q \cdot \bm{v}^q )), \nonumber
\end{align}
where $\bm{v}^q, \bm{v}^a \in \mathbb{R}^{D}$ are the composed vectors of questions and answers respectively, $\bm{b}^a_i$ is the $i$-th word representation in $\bm{B}^a = [\bm{b}^a_1, ..., \bm{b}^a_L]$, $\alpha_i$ is the attention weight for the $i$-th word which is calculated by a softmax function, $\bm{W}^q, \bm{W}^a \in \mathbb{R}^{M \times D}$, $\bm{m} \in \mathbb{R}^{M}$ are attention parameters with $M$ being the hidden size of attention. 

The above formulation is for training. During test procedure, we just need to replace $\bm{B}^a$ by $\bm{\mathcal{B}}^a$.

\subsection{Similarity Layer and Loss Function}
The \emph{similarity layer} measures the similarity between question-answer pairs based on their vector representations $\bm{v}^q$ and $\bm{v}^a$. Here, we choose cosine function as the similarity function, which is usually adopted in answer selection tasks:
\begin{equation}
s = cos(\bm{v}^q, \bm{v}^a), \nonumber
\end{equation}
where $s \in \mathbb{R}$ is the similarity between question $q$ and answer $a$.

Based on the similarity between questions and answers, we can define the loss function. The most commonly used loss function for ranking is the triplet-based hinge loss~\cite{tan2016lstm, tran2018multihop}. To combine the hinge loss and the binary constraint in hashing together, we can get the following optimization problem:
\begin{align}
\min\limits_{\theta, \bm{\mathcal{B}}^a} \mathcal{J} &= \sum_{i=1}^N [ \mathcal{J}^m_i + \delta \cdot \mathcal{J}^c_i] \nonumber \\
&= \sum_{i=1}^N [ max(0, 0.1 - s^+_i + s^-_i) + \delta \cdot \sum_{a \in \{a^+_i, a^-_i\}} ||\bm{B}^a - \bm{\mathcal{B}}^a||_F^2 ], \nonumber
\end{align}
where $\mathcal{J}^m_i = max(0, 0.1 - s^+_i + s^-_i)$ is the hinge loss for the $i$-th example.  $\delta$ is the coefficient of the binary constraint $\mathcal{J}^c_i$. $s^+_i = cos(\bm{v}^{q_i}, \bm{v}^{a_i^+})$ is the similarity between a question $q_i$ and a positive answer $a_i^+$ corresponding to $q_i$. $s^-_i = cos(\bm{v}^{q_i}, \bm{v}^{a_i^-})$ is the similarity between $q_i$ and a randomly selected negative answer $a_i^-$. $N$ is the number of training examples. $\theta$ donates parameters in HAS except $\bm{\mathcal{B}}^a$.

These two sets of parameters $\theta$ and $\bm{\mathcal{B}}^a$ can be optimized alternately~\cite{li2016feature}. More specifically, the $\bm{\mathcal{B}}^a$ can be optimized as follows when $\theta$ is fixed:
\begin{equation}
\bm{\mathcal{B}}^a = sgn(\bm{B}^a). \nonumber
\end{equation}
And $\theta$ can be updated by utilizing back propagation~(BP) when $\bm{\mathcal{B}}^a$ is fixed.

\section{Experiment} \label{sec:exp}
\subsection{Datasets}
We evaluate HAS on three popular answer selection datasets. The statistics about the datasets are presented in Table~\ref{datasets}.
\begin{table}[th]
	\vspace{-10pt}
	\small
	\centering
	\caption{Statistics of the datasets. \label{datasets}}
	\vspace{-5pt}
	\begin{tabular}{lccc}
		\hline
		& insuranceQA & yahooQA & wikiQA \\
		\hline
		number of questions (Train) & 12887 & 50112 & 873\\
		number of questions (Dev)   & 1000  & 6289  & 126\\
		number of questions (Test1) & 1800  & 6283  & 243\\
		number of questions (Test2) & 1800  & ---  & --- \\
		\hline
		number of answers per question & 500 & 5 & 9 \\
		\hline
	\end{tabular}
	\vspace{-10pt}
\end{table}

\textbf{insuraceQA} is a FAQ dataset from insurance domain released by~\cite{feng2015applying}. We use the first version of this dataset, which has been widely used in existing works~\cite{tan2016lstm,wang2016inner,tan2016improved,deng2018knowledge,tran2018multihop}. This dataset has already been partitioned into four subsets, Train, Dev, Test1 and Test2. The total size of candidate answers is 24981. To reduce the complexity, the dataset has provided a candidate set of 500 answers for each question, including positive and negative answers. There is more than one positive answer to some questions. As in existing works~\cite{feng2015applying,tran2018multihop,deng2018knowledge}, we adopt Precision@1~(P@1) as the evaluation metric.

\textbf{yahooQA}~\footnote{\url{https://webscope.sandbox.yahoo.com/catalog.php?datatype=l&guccounter=1}} is a large CQA corpus collected from Yahoo! Answers. We adopt the dataset splits as those in~\cite{tay2017learning,tay2018cross,deng2018knowledge} for fair comparison. Questions and answers are filtered by their length, and only sentences with length among the range of 5 - 50 are preserved. The candidate set for each question is five, in which only one answer is positive. The other four negative answers are sampled from the top $1000$ hits using Lucene search for each question. As in existing works~\cite{tay2017learning,tay2018cross,deng2018knowledge},  P@1 and Mean Reciprocal Rank~(MRR) are adopted as evaluation metrics.

\textbf{wikiQA}~\cite{yang2015wikiqa} is another benchmark for open-domain answer selection. The questions of wikiQA are factual questions which are collected from Bing search logs. Each question is linked to a Wikipedia page, and the sentences in the summary section are collected as the candidate answers. The size of candidate set for each question is different and there may be more than one positive answer to some questions. We filter out the questions which has no positive answers as previous works~\cite{yang2015wikiqa,deng2018knowledge,wang2016inner}. Mean Average Precision~(MAP) and MRR are adopted as evaluation metrics as in existing works.

\subsection{Hyperparameters and Baselines}
We use base BERT as the encoder in our experiments. Large BERT may have better performance, but the \emph{encoding layer} is not the focus of this paper. More specifically, the embedding size $E$ and output dimension $D$ of BERT are $768$. The probability of dropout is $0.1$. Weight decay coefficient is $0.01$. Batch size is $64$ for yahooQA, and $32$ for insuranceQA and wikiQA. The attention hidden size $M$ for insuranceQA is $768$. $M$ is $128$ for yahooQA and wikiQA. Learning rate is $5e^{-6}$ for all models. Training epochs are chosen to achieve the best result on a validation set. All reported results are the average of $5$ runnings.

There are also two other important parameters, $\beta$ in $tanh(\beta x)$ and the coefficient $\delta$ of the binary constraint. $\beta$ is tuned among $\{1, 2, 5,$$ 10, 20\}$, and $\delta$ is tuned among $\{0, 1e^{-7}, 1e^{-6}, 1e^{-5}, 1e^{-4}\}$. 

The state-of-the-art baselines on three datasets are different. Hence, we adopt different baselines for comparison on different datasets according to previous works. Baselines using single model without extra knowledge include: CNN, CNN with GESD~\cite{feng2015applying}, QA-LSTM~\cite{tan2016lstm}, \mbox{AP-LSTM}~\cite{tran2018multihop}, \mbox{Multihop-Sequential-LSTM}~\cite{tran2018multihop}, IARNN-GATE~\cite{wang2016inner}, NTN-LSTM, HD-LSTM~\cite{tay2017learning}, HyperQA~\cite{tay2018hyperbolic}, AP-CNN~\cite{santos2016attentive}, AP-BiLSTM~\cite{santos2016attentive}, CTRN~\cite{tay2018cross}, CA-RNN~\cite{chen2018rnn}, RNN-POA~\cite{chen2017enhancing}, MULT~\cite{wang2017a}, MV-FNN~\cite{sha2018a}. Single models with external knowledge include: KAN~\cite{deng2018knowledge}. Ensemble models include: LRXNET~\cite{narayan2018document}, SUM$_{BASE,PTK}$~\cite{tymoshenko2018cross}.

Because HAS adopts BERT as encoder, we also construct two BERT-based baselines for comparison. \emph{BERT-pooling} is a model in which both questions and answers are composed into vectors by pooling. \emph{BERT-attention} is a model which adopts attention as the composition module. Both BERT-pooling and BERT-attention use BERT as the encoder, and hashing is not adopted in them. 

\subsection{Experimental Results}
\begin{wraptable}{r}{0.65\textwidth}
	\vspace{-25pt}
	\begin{center}
	\small
	\caption{Results on insuranceQA. The results of models marked with $\star$ are reported from~\cite{tran2018multihop}. Other results marked with $\diamond$ are reported from their original paper. P@1 is adopted as evaluation metric as previous works. `our impl.' denotes our implementation. \label{exp2}}
	\vspace{-10pt}
	\begin{tabular}{lcc}
		\hline
		Model & P@1~(Test1) & P@1~(Test2) \\
		\hline
		CNN $\star$ & 62.80 & 59.20 \\
		CNN with GESD $\star$ & 65.30 & 61.00 \\
		QA-LSTM (our impl.) & 66.08 & 62.63 \\
		AP-LSTM $\star$ & 69.00 & 64.80 \\
		IARNN-GATE $\star$ & 70.10 & 62.80 \\
		Multihop-Sequential-LSTM $\star$ & 70.50 & 66.90 \\
		AP-CNN $\diamond$ & 69.80 & 66.30 \\
		AP-BiLSTM $\diamond$& 71.70 & 66.40 \\
		MULT $\diamond$& 75.20 & 73.40 \\
		KAN~(Tgt-Only) $\diamond$& 71.50 & 68.80 \\
		KAN $\diamond$& 75.20 & 72.50 \\
		\hline
		HAS & \textbf{76.38} & \textbf{73.71} \\
		\hline
	\end{tabular}
	\end{center}
	\vspace{-15pt}
\end{wraptable}
\paragraph{Results on insuranceQA} We compare HAS with baselines on \mbox{insuranceQA} dataset. The results are shown in Table~\ref{exp2}. MULT~\cite{wang2017a} and KAN~\cite{deng2018knowledge} are two strong baselines which represent the state-of-the-art results on this dataset. Here, KAN adopts external knowledge for performance improvement. KAN~(Tgt-Only) denotes the KAN variant without external knowledge. We can find that HAS outperforms all the baselines, which proves the effectiveness of HAS.

\paragraph{Results on yahooQA} We also evaluate HAS and baselines on yahooQA. Table~\ref{exp3} shows the results. KAN~\cite{deng2018knowledge}, which utilizes external knowledge, is the state-of-the-art model on this dataset. HAS outperforms all baselines except KAN. The performance gain of KAN mainly owes to the external knowledge, by pre-training on a source QA dataset \mbox{SQuAD-T}. Please note that HAS does not adopt external QA dataset for pre-training. HAS can outperform the target-only version of KAN, denoted as KAN~(Tgt-Only), which is only trained on \mbox{yahooQA} without \mbox{SQuAD-T}. Once again, the result on \mbox{yahooQA} verifies the effectiveness of HAS.

\begin{table}[t]
  \begin{minipage}{0.48\textwidth}
      \centering
      \small
      \caption{Results on yahooQA. The results of models marked with $\star$ are reported from~\cite{tay2018cross}. Other results marked with $\diamond$ are reported from their original paper. P@1 and MRR are adopted as evaluation metrics as previous works. \label{exp3}}
      \vspace{-6pt}
      \begin{tabular}{lcc}
		\hline
		Model & P@1 & MRR \\
		\hline
		Random Guess & 20.00 & 45.86 \\
		NTN-LSTM $\star$ & 54.50 & 73.10 \\
		HD-LSTM $\star$ & 55.70 & 73.50 \\
		AP-CNN $\star$ & 56.00 & 72.60 \\
		AP-BiLSTM $\star$ & 56.80 & 73.10 \\
		CTRN $\star$ & 60.10 & 75.50 \\
		HyperQA $\diamond$ & 68.30 & 80.10	\\
		KAN~(Tgt-Only) $\diamond$ & 67.20 & 80.30 \\
		KAN $\diamond$& \textbf{74.40} & \textbf{84.00} \\
		\hline
		HAS & 73.89 & 82.10 \\
		\hline
      \end{tabular}
   \end{minipage} \hspace{6pt}
   \begin{minipage}{0.49\textwidth}
     \centering
     \small
     \caption{Results on wikiQA. The results marked with $\diamond$ are reported from their original paper. MAP and MRR are adopted as evaluation metrics as previous works. \label{exp4}}
     \vspace{-6pt}
     \begin{tabular}{lcc}
		\hline
		Model & MAP & MRR \\
		\hline
		AP-CNN  $\diamond$ & 68.86 & 69.57 \\
		AP-BiLSTM $\diamond$& 67.05 & 68.42 \\
		RNN-POA $\diamond$& 72.12 & 73.12 \\
		Multihop-Sequential-LSTM $\diamond$& 72.20 & 73.80 \\
		IARNN-GATE $\diamond$& 72.58 & 73.94 \\
		CA-RNN $\diamond$& 73.58 & 74.50 \\
		MULT $\diamond$& 74.33 & 75.45 \\
		MV-FNN $\diamond$& 74.62 & 75.76 \\
		SUM$_{BASE,PTK}$ $\diamond$& 75.59 & 77.00 \\
		LRXNET $\diamond$& 76.57 & 75.10 \\
		\hline
		HAS & \textbf{81.01} & \textbf{82.22} \\
		\hline
     \end{tabular}
   \end{minipage}
   \vspace{-15pt}
\end{table}

\paragraph{Results on wikiQA} Table~\ref{exp4} shows the result on wikiQA dataset. SUM$_{BASE,PTK}$~\cite{tymoshenko2018cross} and LRXNET~\cite{narayan2018document} are two ensemble models which represent the state-of-the-art results on this dataset. \mbox{HAS} outperforms all the baselines again, which further proves the effectiveness of our HAS.

\paragraph{Comparison with BERT-based Models} We compare HAS with BERT-pooling and BERT-attention on three datasets. As shown in Table~\ref{exp5}, BERT-attention and HAS outperform BERT-pooling on all three datasets, which verifies that question-answer interaction mechanisms have better performance than pooling. Furthermore, we can find that HAS can achieve comparable accuracy as BERT-attention. But BERT-attention has either speed~(time cost) problem or memory cost problem, which will be shown in the following subsection.

\begin{table*}[t]
	\centering
	\small
	\caption{Comparison with BERT-based models. \label{exp5}}
	\begin{tabular}{lcccccc}
		\hline
		& \multicolumn{2}{c}{insuranceQA} & \multicolumn{2}{c}{yahooQA} & \multicolumn{2}{c}{wikiQA} \\
		\hline
		Model          & P@1 (Test1) & P@1 (Test2) & P@1 & MRR & MAP & MRR\\
		\hline
		BERT-pooling   & 74.52 & 71.97 &73.49&81.93&77.22&78.27\\
		BERT-attention & 76.12 & \textbf{74.12} &\textbf{74.78}&\textbf{82.68}&80.65&81.83\\
		\hline
		HAS  		     & \textbf{76.38} & 73.71 &73.89&82.10&\textbf{81.01}&\textbf{82.22}\\
		\hline
	\end{tabular}
	\vspace{-15pt}
\end{table*}

\paragraph{Time Cost and Memory Cost} To further prove the effectiveness of HAS, we compare HAS with baselines on insuranceQA in terms of time cost and memory cost when the model is deployed for prediction. The results are shown in Table~\ref{exp6}. All experiments are run on a Titan XP GPU. BERT-pooling can directly store the vector representations of answers with a low memory cost, which doesn't have the time cost and memory cost problem. But the accuracy of BERT-pooling is much lower than BERT-attention and HAS. BERT-attention~(recal.) denotes a BERT-attention variant with recalculation, and BERT-attention~(store) denotes a BERT-attention variant which stores the matrix representations of answers in memory. BERT-attention~(recal.) does not need to store the matrix representations of answers in memory, and BERT-attention~(store) does not need recalculation.  The time cost is $4.19$ seconds per question for BERT-attention~(recal.), which is $15$ times slower than HAS. Although BERT-attention~(store) has low time cost as that of HAS, the memory cost of it is $14.29$ GB, which is $32$ times larger than that of HAS. 

We also compare HAS with other baselines in existing works. KAN and MULT do question-answer interaction before \emph{encoding layer} or during \emph{encoding layer}, and the outputs of the \emph{encoding layer} for an answer are different for different questions. Thus, these two models cannot store representations for reusing. We compare HAS with Multihop-Sequential-LSTM, AP-CNN, and AP-BiLSTM. The memory cost of these three models is $5.25$ GB, $7.44$ GB, and $5.25$ GB, respectively, which are $11.75$, $16.67$, $11.75$ times larger than that of HAS. Other baselines are not adopted for comparison, but almost all baselines with question-answer interaction mechanisms have either time cost problem or memory cost problem as that in BERT-attention.

We can find that our HAS is fast with a low memory cost, which also makes HAS have promising potential for embedded or mobile applications.

\begin{table*}[htb]
	\vspace{-10pt}
	\centering
	\small
	\caption{Comparison of accuracy, time cost and memory cost on insuranceQA. Each question has $500$ candidate answers. ``Memory Cost $\diamond$ '' is the memory cost for storing representations of answers. \label{exp6}}
	\begin{tabular}{lcccc}
		\hline
		Model                   & P@1  (Test1)   & P@1 (Test2) & Time Cost per Question & Memory Cost $\diamond$\\
		\hline
		BERT-pooling     		& 74.52   & 71.97 & 0.03s & 0.07 GB \\
		BERT-attention~(recal.) & 76.12   & 74.12 & 4.19s & 0.02 GB \\
		BERT-attention~(store)  & 76.12   & 74.12 & 0.28s & 14.29 GB\\
		\hline
		HAS  					& 76.38   & 73.71 & 0.28s & 0.45 GB\\
		\hline
		Multihop-Sequential-LSTM& 70.50   & 66.90 & 0.13s & 5.25 GB\\
		AP-CNN                  & 69.80   & 66.30 & ---  & 7.44 GB\\
		AP-BiLSTM               & 71.70   & 66.40 & ---  & 5.25 GB\\
		\hline
	\end{tabular}
	\vspace{-15pt}
\end{table*}

\begin{figure}[h!]
	\vspace{-10pt}
	\centering
	\subfigure[Sensitivity of $\beta$ when $\delta=1e^{-6}$]{
		\label{exp:beta-insurance}
		\includegraphics[width=0.35\textwidth]{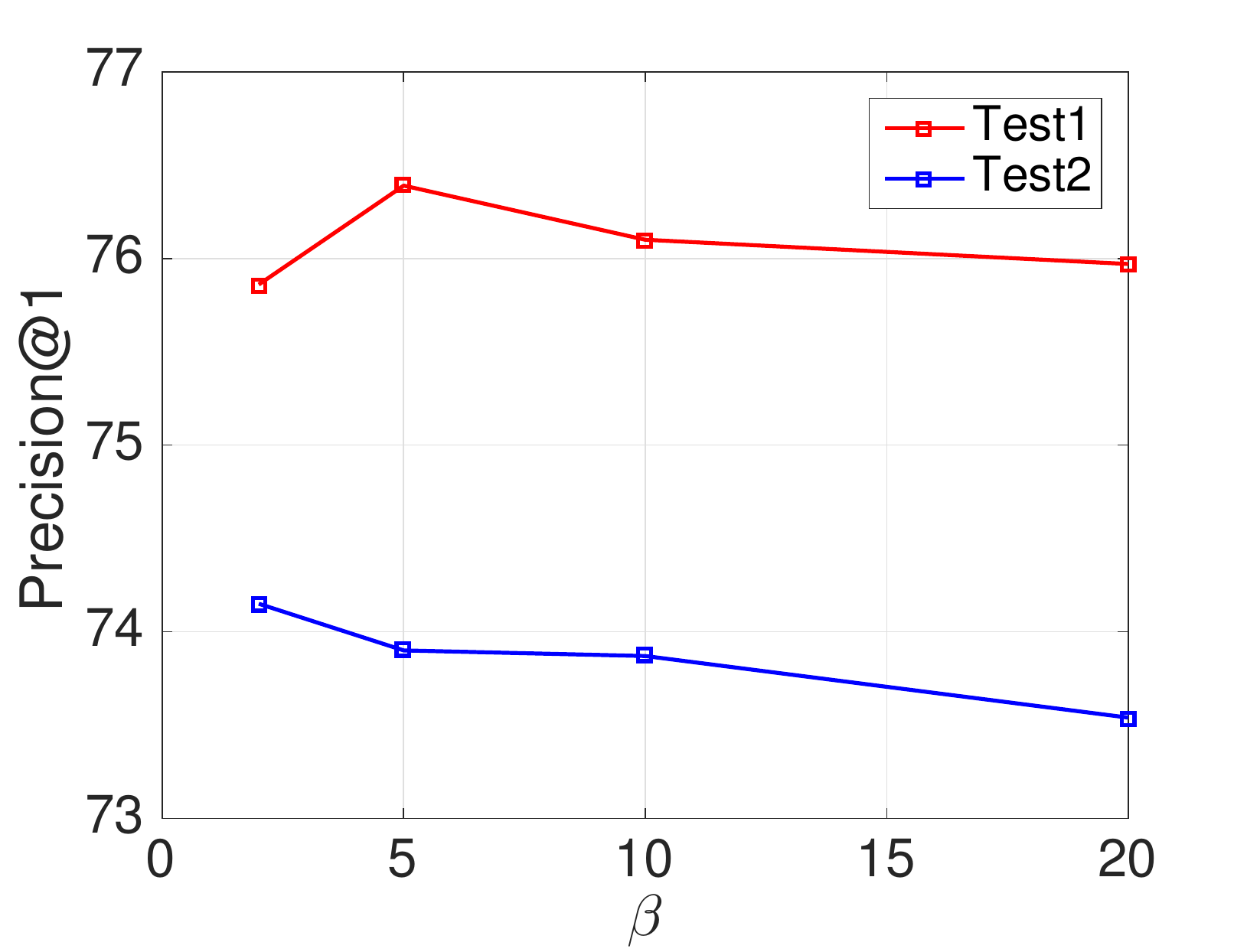}}\hspace{30pt}
	\subfigure[Sensitivity of $\delta$ when $\beta=5$]{
		\label{exp:delta-insurance}
		\includegraphics[width=0.35\textwidth]{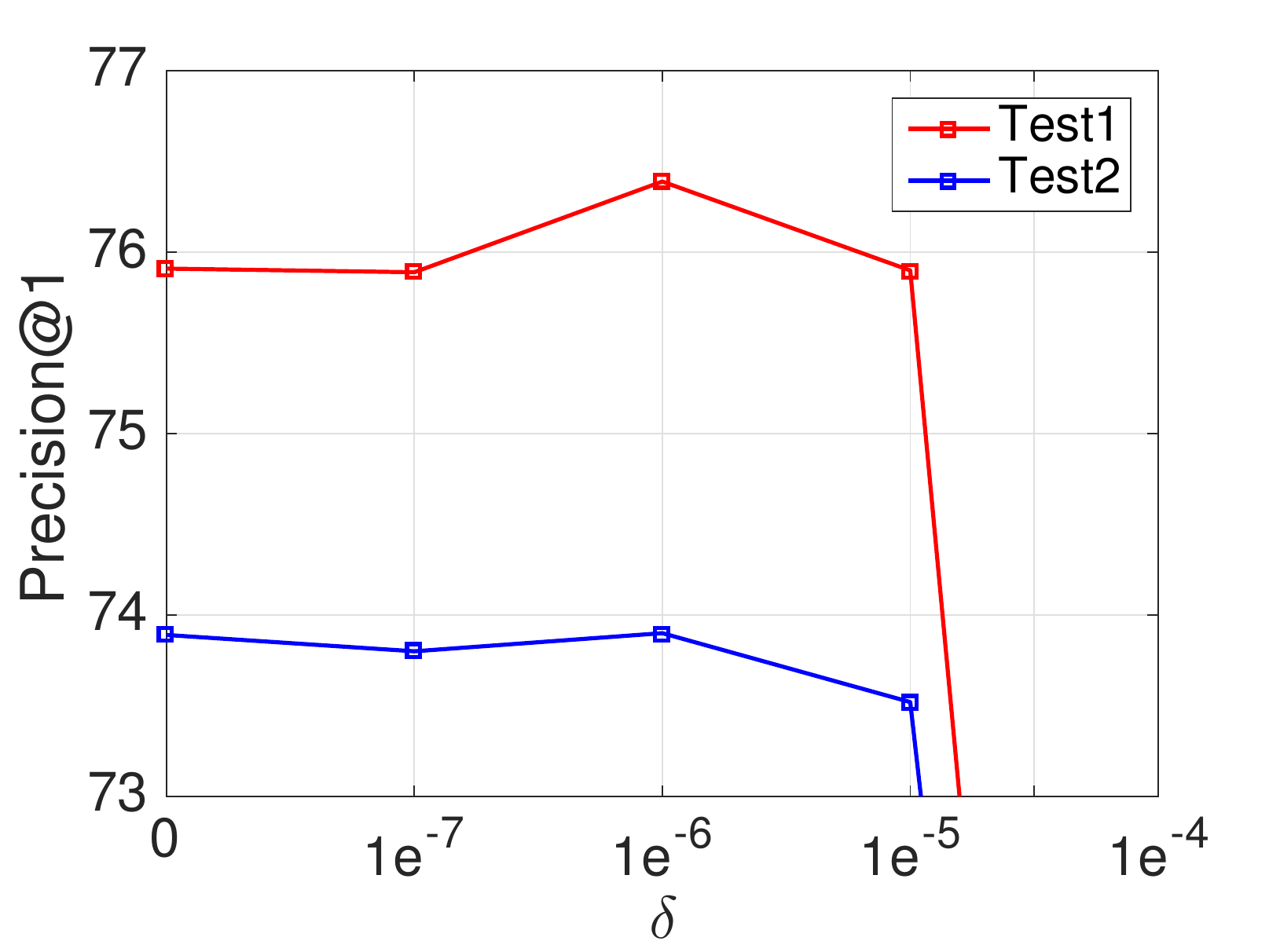}}
	\vspace{-5pt}
	\caption{Sensitivity analysis on insuranceQA.}
\end{figure}

\begin{figure}[h!]
	\vspace{-20pt}
	\centering
	\subfigure[Sensitivity of $\beta$ when $\delta=1e^{-6}$]{
		\label{exp:beta}
		\includegraphics[width=0.35\textwidth]{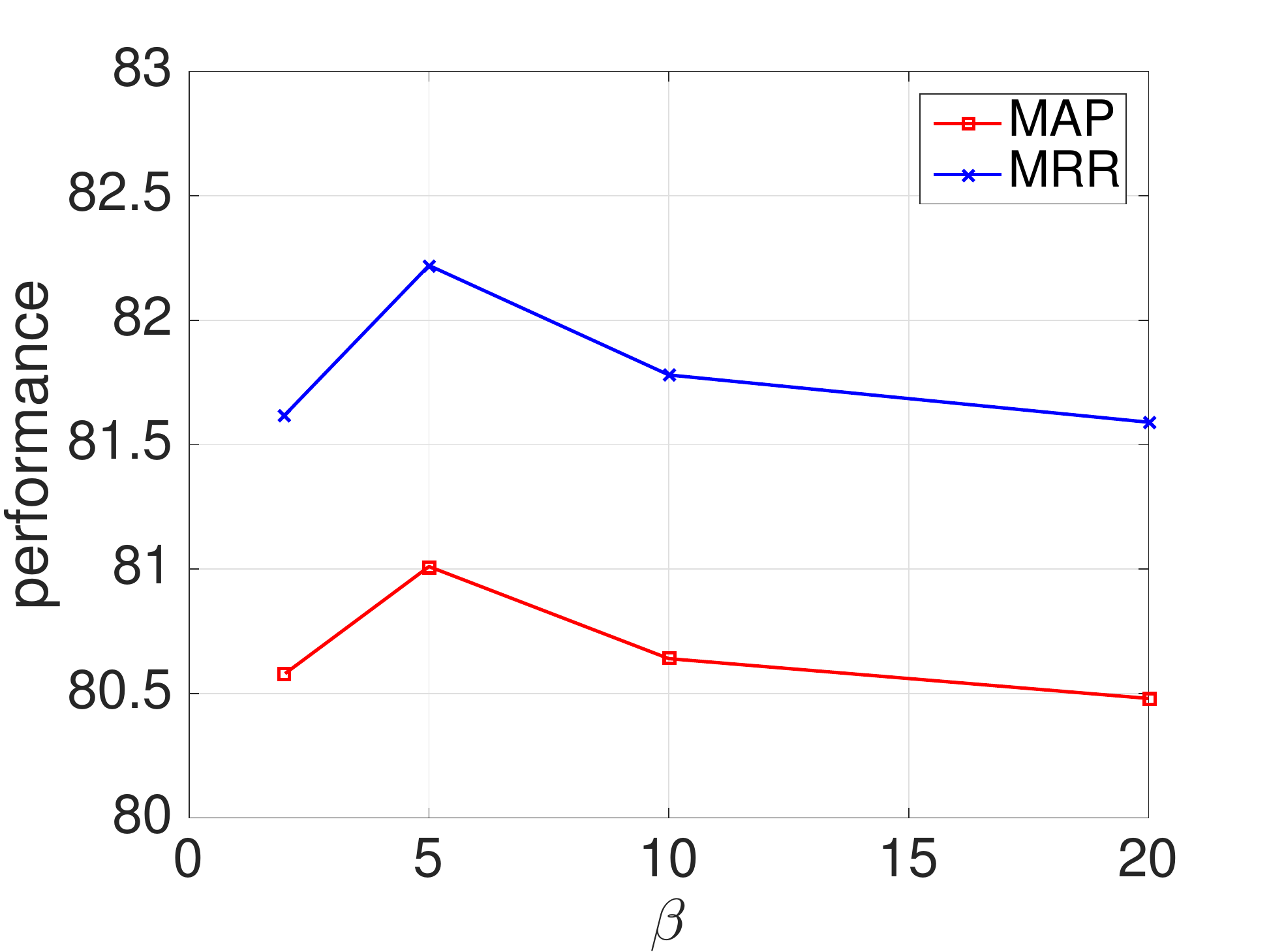}}\hspace{30pt}
	\subfigure[Sensitivity of $\delta$ when $\beta=5$]{
		\label{exp:delta}
		\includegraphics[width=0.35\textwidth]{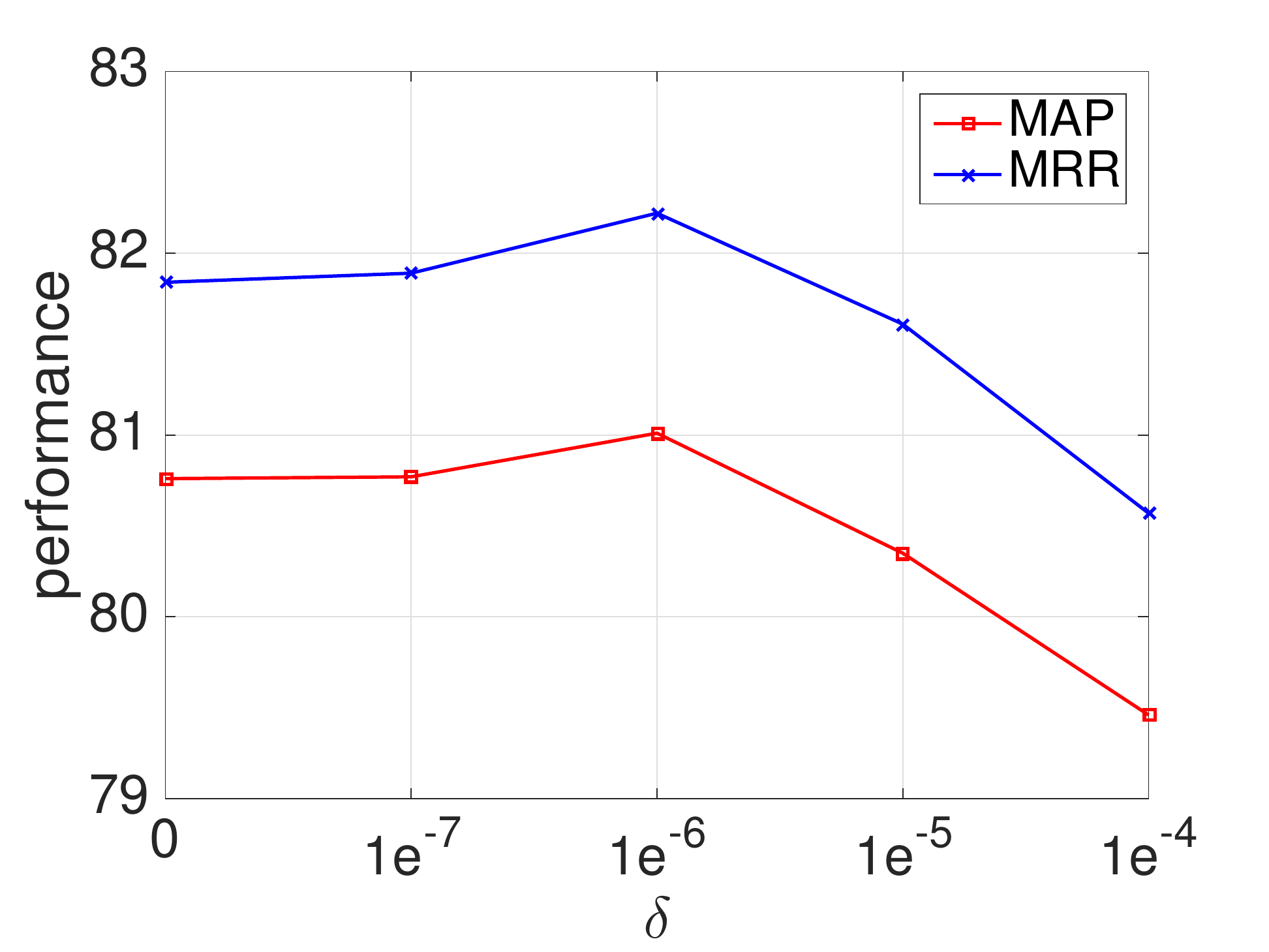}}
	\vspace{-5pt}
	\caption{Sensitivity analysis on wikiQA.}
	\vspace{-10pt}
\end{figure}

\paragraph{Sensitivity Analysis of $\delta$ and $\beta$} In this section, we study the sensitivity of the two important hyper-parameters in HAS, which are the coefficient $\delta$ of $\mathcal{J}^c$ and the value of $\beta$ in $tanh(\beta x)$. We design a sensitivity study of these two hyper-parameters on insuranceQA and wikiQA. As shown in Figure~\ref{exp:beta-insurance} and Figure~\ref{exp:beta}, the performance can be improved by increasing $\beta$ to $5$. We can find that HAS is not sensitive to $\beta$ in the range of [5, 10]. When $\beta$ is fixed to $5$, the performance of different choices of $\delta$ is shown in Figure~\ref{exp:delta-insurance} and Figure~\ref{exp:delta}.  We can find that HAS is not sensitive to $\delta$ in the range of [$1e^{-7}, 1e^{-5}$]. 

  

\paragraph{A Case Study} Table~\ref{casestudy} gives a hard example in wikiQA. Only the first answer is positive. The key to distinguish these two answers is the word ``popular'' in the question. Answer1 is selected by HAS, while Answer2 is selected by QA-LSTM which is a RNN-based model trained from scratch. From this case, we can find that the common knowledge~(the relation between ``popular'' and ``TV'' shows, ``audiences'') plays a crucial role in some situations. By using the hashing technique, HAS can adopt complex encoders like BERT and GPT-2 to encode abundant common knowledge in the model, with a low time cost and low memory cost.
\begin{table}[ht]
	\vspace{-5pt}
	\centering
	\small
	\caption{An example in wikiQA, which is hard to answer without any common knowledge. \label{casestudy}}
	\vspace{-5pt}
	\begin{tabular}{ll}
		\hline
		Question  & What is a popular people meter ? \\
		\hline
		Answer1~(positive)& A people meter is an audience measurement tool used to measure\\
		               & viewing habits of TV and cable audiences. \\
		\hline
		Answer2~(negative)& The People Meter is a 'box', about the size of a paperback book . \\
		\hline
	\end{tabular}
	\vspace{-15pt}
\end{table}

\section{Conclusion}
In this paper, we propose a novel answer selection method called \underline{h}ashing based \underline{a}nswer \underline{s}election~(\mbox{HAS}). HAS adopts hashing to learn binary matrix representations for answers, which can dramatically reduce memory cost for storing the matrix outputs of encoders in answer selection. When deployed for prediction, HAS is fast with a low memory cost. This is particularly meaningful when the model needs to be deployed at embedded or mobile systems. Experimental results on three popular datasets show that HAS can outperform existing methods to achieve state-of-the-art performance.

HAS is flexible to integrate other encoders and question-answer interaction mechanisms, which will be pursued in our future work. 

\small
\bibliography{template_page}
\bibliographystyle{abbrvnat}

\end{document}